%% file: paper2593.tex
\RequirePackage{amsmath}

\documentclass[runningheads]{llncs}
\usepackage{hyperref}       
\usepackage{url}            
\usepackage{cite, amssymb}
\usepackage{xcolor}

\usepackage[T1]{fontenc}
\usepackage{graphicx}
\begin{document}
\input{notations.tex}
%
\title{\ssmmodel: Reimagining Statistical Shape Models from Images with Radial Basis Functions}
\titlerunning{\ssmmodel: Statistical Shape Models from Images}
%
%
\author{Hong Xu\inst{1} \and Shireen Y. Elhabian\inst{1}}
%
\authorrunning{H. Xu \and S. Y. Elhabian}
%
\institute{Scientific Computing and Imaging Institute, Kahlert School of Computing, \\University of Utah, Salt Lake City, UT, USA \\
\url{http://www.sci.utah.edu}, \url{https://www.cs.utah.edu} \\
\email{\{hxu,shireen\}@sci.utah.edu} 
}
%
\maketitle              

\input{abstract.tex}
\input{introduction.tex}

\input{methods.tex}
\input{results.tex}
\input{conclusion.tex}

%
%
%
%
\bibliographystyle{splncs04}
\bibliography{references}

\input{supplementary.tex}

\end{document}

%% file: notations.tex

\newcommand{\ssmmodel}{Image2SSM}

\newcommand\ShapeRepresentation{Radial-Basis-Function Shape}
\newcommand\shaperepresentation{radial basis function shape}
\newcommand\srepresentation{RBF-shape}
\newcommand\samploss{narrow band loss}

\newcommand\Real{\mathbb{R}}
\newcommand\diff{\mathbf{d}}
\newcommand\aP{\mathbf{\mathcal{P}}}
\newcommand\aN{\mathbf{\mathcal{N}}}
\newcommand\aD{\mathbf{\mathcal{D}}}
\newcommand\bN{\mathbf{N}}


\def\x{{\mathbf x}}
\def\L{{\cal L}}

\newcommand{\var}{{\rm var}}
\newcommand{\Tr}{^{\rm T}}
\newcommand{\vtrans}[2]{{#1}^{(#2)}}
\newcommand{\kron}{\otimes}
\newcommand{\schur}[2]{({#1} | {#2})}
\newcommand{\schurdet}[2]{\left| ({#1} | {#2}) \right|}
\newcommand{\had}{\circ}
\newcommand{\diag}{{\rm diag}}
\newcommand{\invdiag}{\diag^{-1}}
\newcommand{\rank}{{\rm rank}}
\newcommand{\nullsp}{{\rm null}}
\newcommand{\tr}{{\rm tr}}
\newcommand{\vech}{{\rm vech}}
\renewcommand{\det}[1]{\left| #1 \right|}
\newcommand{\pdet}[1]{\left| #1 \right|_{+}}
\newcommand{\pinv}[1]{#1^{+}}
\newcommand{\erf}{{\rm erf}}
\newcommand{\hypergeom}[2]{{}_{#1}F_{#2}}

\renewcommand{\a}{{\bf a}}
\renewcommand{\b}{{\bf b}}
\renewcommand{\c}{{\bf c}}
\renewcommand{\d}{{\rm d}}  
\newcommand{\e}{{\bf e}}
\newcommand{\f}{{\bf f}}
\newcommand{\g}{{\bf g}}
\newcommand{\h}{{\bf h}}
\renewcommand{\k}{{\bf k}}
\newcommand{\m}{{\bf m}}
\newcommand{\mb}{{\bf m}}
\newcommand{\n}{{\bf n}}
\renewcommand{\o}{{\bf o}}
\newcommand{\p}{{\bf p}}
\newcommand{\q}{{\bf q}}
\renewcommand{\r}{{\bf r}}
\newcommand{\s}{{\bf s}}
\renewcommand{\t}{{\bf t}}
\renewcommand{\u}{{\bf u}}
\renewcommand{\v}{{\bf v}}
\newcommand{\w}{{\bf w}}
\newcommand{\y}{{\bf y}}
\newcommand{\z}{{\bf z}}
\newcommand{\A}{{\bf A}}
\newcommand{\B}{{\bf B}}
\newcommand{\C}{{\bf C}}
\newcommand{\D}{{\bf D}}
\newcommand{\E}{{\bf E}}
\newcommand{\F}{{\bf F}}
\newcommand{\G}{{\bf G}}
\renewcommand{\H}{{\bf H}}
\newcommand{\I}{{\bf I}}
\newcommand{\J}{{\bf J}}
\newcommand{\K}{{\bf K}}
\renewcommand{\L}{{\bf L}}
\newcommand{\M}{{\bf M}}
\newcommand{\N}{\mathcal{N}}  
\renewcommand{\O}{{\bf O}}
\renewcommand{\P}{{\bf P}}
\newcommand{\Q}{{\bf Q}}
\newcommand{\R}{{\bf R}}
\renewcommand{\S}{{\bf S}}
\newcommand{\T}{{\bf T}}
\newcommand{\U}{{\bf U}}
\newcommand{\V}{{\bf V}}
\newcommand{\W}{{\bf W}}
\newcommand{\X}{{\bf X}}
\newcommand{\Y}{{\bf Y}}
\newcommand{\Z}{{\bf Z}}

\newcommand{\bfLambda}{\boldsymbol{\Lambda}}

\newcommand{\bsigma}{\boldsymbol{\sigma}}
\newcommand{\balpha}{\boldsymbol{\alpha}}
\newcommand{\bpsi}{\boldsymbol{\psi}}
\newcommand{\bphi}{\boldsymbol{\phi}}
\newcommand{\boldeta}{\boldsymbol{\eta}}
\newcommand{\Beta}{\boldsymbol{\eta}}
\newcommand{\btau}{\boldsymbol{\tau}}
\newcommand{\bvarphi}{\boldsymbol{\varphi}}
\newcommand{\bzeta}{\boldsymbol{\zeta}}

\newcommand{\blambda}{\boldsymbol{\lambda}}
\newcommand{\bLambda}{\mathbf{\Lambda}}
\newcommand{\bOmega}{\mathbf{\Omega}}
\newcommand{\bomega}{\mathbf{\omega}}
\newcommand{\bPi}{\mathbf{\Pi}}

\newcommand{\btheta}{\boldsymbol{\theta}}
\newcommand{\bpi}{\boldsymbol{\pi}}
\newcommand{\bxi}{\boldsymbol{\xi}}
\newcommand{\bSigma}{\boldsymbol{\Sigma}}

\newcommand{\bgamma}{\boldsymbol{\gamma}}
\newcommand{\bGamma}{\mathbf{\Gamma}}

\newcommand{\bmu}{\boldsymbol{\mu}}
\newcommand{\1}{{\bf 1}}
\newcommand{\0}{{\bf 0}}

\newcommand{\bs}{\backslash}
\newcommand{\ben}{\begin{enumerate}}
\newcommand{\een}{\end{enumerate}}

 \newcommand{\notS}{{\backslash S}}
 \newcommand{\nots}{{\backslash s}}
 \newcommand{\noti}{{\backslash i}}
 \newcommand{\notj}{{\backslash j}}
 \newcommand{\nott}{\backslash t}
 \newcommand{\notone}{{\backslash 1}}
 \newcommand{\nottp}{\backslash t+1}

\newcommand{\notk}{{^{\backslash k}}}
\newcommand{\notij}{{^{\backslash i,j}}}
\newcommand{\notg}{{^{\backslash g}}}
\newcommand{\wnoti}{{_{\w}^{\backslash i}}}
\newcommand{\wnotg}{{_{\w}^{\backslash g}}}
\newcommand{\vnotij}{{_{\v}^{\backslash i,j}}}
\newcommand{\vnotg}{{_{\v}^{\backslash g}}}
\newcommand{\half}{\frac{1}{2}}
\newcommand{\msgb}{m_{t \leftarrow t+1}}
\newcommand{\msgf}{m_{t \rightarrow t+1}}
\newcommand{\msgfp}{m_{t-1 \rightarrow t}}

\newcommand{\proj}[1]{{\rm proj}\negmedspace\left[#1\right]}
\newcommand{\argmin}{\operatornamewithlimits{argmin}}
\newcommand{\argmax}{\operatornamewithlimits{argmax}}

\newcommand{\dif}{\mathrm{d}}
\newcommand{\abs}[1]{\lvert#1\rvert}
\newcommand{\norm}[1]{\lVert#1\rVert}

\newcommand{\ie}{{{i.e.,}}\xspace}
\newcommand{\eg}{{{\em e.g.,}}\xspace}
\newcommand{\EE}{\mathbb{E}}
\newcommand{\VV}{\mathbb{V}}
\newcommand{\sbr}[1]{\left[#1\right]}
\newcommand{\rbr}[1]{\left(#1\right)}
\newcommand{\cmt}[1]{}

%% file: abstract.tex
\begin{abstract}

Statistical shape modeling (SSM) is an essential tool for analyzing variations in anatomical morphology. In a typical SSM pipeline, 3D anatomical images, gone through segmentation and rigid registration, are represented using lower-dimensional shape features, on which statistical analysis can be performed. Various methods for constructing compact shape representations have been proposed, but they involve laborious and costly steps. We propose \ssmmodel, a novel deep-learning-based approach for SSM that leverages image-segmentation pairs to learn a radial-basis-function (RBF)-based representation of shapes directly from images. This RBF-based shape representation offers a rich self-supervised signal for the network to estimate a continuous, yet compact representation of the underlying surface that can adapt to complex geometries in a data-driven manner. \ssmmodel\ can characterize populations of biological structures of interest by constructing statistical landmark-based shape models of ensembles of anatomical shapes while requiring minimal parameter tuning and no user assistance. Once trained, \ssmmodel~can be used to infer low-dimensional shape representations from new unsegmented images, paving the way toward scalable approaches for SSM, especially when dealing with large cohorts. Experiments on synthetic and real datasets show the efficacy of the proposed method compared to the state-of-art correspondence-based method for SSM.


\keywords{Statistical Shape Modeling  \and Deep Learning \and Radial Basis Function Interpolation \and Polyharmonic Splines.}
\end{abstract}

%% file: introduction.tex
\section{Introduction}
Statistical Shape Modeling (SSM) or morphological analysis, is a widespread tool used to quantify anatomical shape variation given a population of segmented 3D anatomies. Quantifying such subtle shape differences has been crucial in providing individualized treatments in medical procedures, detecting morphological pathologies, and advancing the understanding of different diseases \cite{atkins2017quantitative, bhalodia2020quantifying, harris2013cam, LenzTalocruralJoint, VANBUURENHipOsteoarthritis, cates2014computational,merle2014many,merle2019high,carriere2014apathy,bruse2016statistical,zachow2015computational,zadpoor2015patient}. 

The two principal shape representations for building SSMs and performing subsequent statistical analyses are \textit{deformation fields} and \textit{landmarks}. Deformation fields encode \textit{implicit} transformations between cohort samples and a pre-defined (or learned) atlas. In contrast, landmarks are \textit{explicit} points spread on shape surfaces that correspond across the population \cite{sarkalkan2014statistical, thompson1942growth}. 
Landmark-based representations have been used extensively due to their simplicity, computational efficiency, and interpretability for statistical analyses \cite{sarkalkan2014statistical,zachow2015computational}. 
Some applications use manually defined landmarks, however, this is labor-intensive, not reproducible, and requires domain expertise (e.g., radiologists).
Computational methods (e.g., minimum description length -- MDL \cite{davies2002minimum}, particle-based shape modeling -- PSM \cite{cates2017shapeworks, cates2007shape}, and frameworks based on Large Deformation Diffeomorphic Metric Mapping \cite{durrleman2014morphometry}) for automatically placing dense \textit{correspondence points}, aka \textit{point distribution models} (PDMs), have shifted the SSM field to data-driven characterization of population-level variabilities that is objective, reproducible, and scalable. 
%
However, this efficiency suffers when intricate shape surfaces require thousands of points representing localized, convoluted shape features that may live between landmarks. Furthermore, existing methods for landmark-based SSM must go through laborious and computationally expensive steps that require anatomical and technical expertise, starting from anatomy segmentation, shape data preprocessing, and correspondence optimization, to generate PDMs from 3D images. Existing methods (e.g., \cite{bhalodia2018deepssm, uncertaindeepssm, bhalodia2021deepssm, adams2022images, Parisot2018, tothova2020probabilistic}) have been able to use deep learning to assuage the arduous process of building a PDM but still require the construction of PDMs (e.g., using a computational method such as PSM \cite{cates2017shapeworks, cates2007shape}) to supervise its learning task, making these deep learning based methods restricted and biased towards the shape statistics captured by the SSM method that is used to construct their training data. 

To address the shortcomings of existing models, we propose \ssmmodel, a novel deep-learning-based approach for SSM directly from images that, given pairs of images and segmentations, can produce a statistical shape model using an implicit, continuous surface representation. 
Once trained, \ssmmodel~can produce PDMs of new images without the need for anatomy segmentations. 
Unlike existing deep learning-based methods for SSM from images, \ssmmodel~only requires image-segmentation pairs and alleviates the need for constructing PDM to supervise learning shape statistics from images.
\ssmmodel~leverages an implicit, radial basis function (RBF)-based, representation of shapes to construct a self-supervised training signal by tasking the network to estimate a sparse set of control points and their respective suface normals that best approximate the underlying surface in the RBF sense. 
%
%
This novel application of RBFs to build SSMs allows statistical analyses on representative points/landmarks, their surface normals, and the shape surfaces themselves due to its compact, informative, yet comprehensive nature. 
Combined with deep networks to directly learn such a representation from images, this method ushers a next step towards fully end-to-end SSM frameworks that can build better and less restrictive low-dimensional shape representations more conducive to SSM analysis. In summary, the proposed method for SSM has the following strengths. 

\begin{itemize}
  \item Using a continuous, but compact surface representation instead of only landmarks that allows performing analyses on points, normals, and surfaces alike. 
  \item The RBF shape representation can adapt to the underlying surface geometry, spreading more landmarks over the more complex surface regions. 
  \item A deep learning approach that bypasses any conventional correspondence optimization to construct training data for supervision, requiring virtually no hyperparameter tuning or preprocessing steps. 
  \item This method uses accelerated computational resources to perform training and outperforms existing deep learning based methods that constructs PDMs from unsegmented images. 
\end{itemize}

%% file: methods.tex
\section{Methods}


\ssmmodel\ is a deep learning method that learns to build an SSM for an anatomical structure of interest directly from unsegmented images. It is trained on a population of $I-$3D images $\mathcal{I} = \{\I_i\}_{i=1}^I$ as input and is supervised by their respective segmentations $\mathcal{S} = \{\S_i\}_{i=1}^I$. \ssmmodel\ learns an RBF-based shape representation, consisting of a set of $J$ control points $\aP = \{\P_i\}_{i=1}^I$, and their surface normals $\aN = \{ \bN_i \}_{i=1}^I$ for each input shape, where the $i-$th shape point distribution model (PDM) is denoted by $\P_i = [\p_{i,1}, \p_{i,2}, \cdots, \p_{i,J}]$, the respective surface normals are $\bN_i = [\n_{i,1}, \n_{i,2}, \cdots, \n_{i,J}]$, and $\p_{i,j}, \n_{i,j} \in \mathbb{R}^3$. 
%
The network is trained end-to-end to minimize a loss that (1) makes the learned control points and their surface normals adhere to the underlying surface, (2) approximates surface normals at each control point to encode a signed distance field to the surface, (3) promotes correspondence of these control points across shapes in the population, and (4) encourages a spread of control points on each surface that adapts to the underlying geometrical complexity.
The learned control points define an anatomical mapping, or a \textit{metric}, among the given shapes that enables quantifying subtle shape differences and performing shape statistics, for example, using principal component analysis (PCA) or other non-linear methods (e.g., \cite{kempfert2020nonlineardimensionalityreduction}). 
More importantly, once trained, \ssmmodel\ can generate PDMs for new unsegmented images, bypassing the conventional SSM workflow of the manual (or semi-automated) segmentation, data preprocessing, and correspondence optimization.  
Furthermore, the continuous, implicit nature of the RBF representation enables extracting a proxy geometry (e.g., surface mesh or signed distance transforms -- SDFs) at an arbitrary resolution that can be rasterized trivially on graphics hardware \cite{Turk1999VariationalIS, rbf_surface1}.

%




In this section, we briefly elaborate on the \srepresentation\ representation, outline the network architecture, motivate the choices and design of the proposed losses, and detail the training protocol of \ssmmodel.


\subsection{Representing shapes using RBFs} \label{rbf_section}

\begin{figure}[!t]
\includegraphics[width= \textwidth]{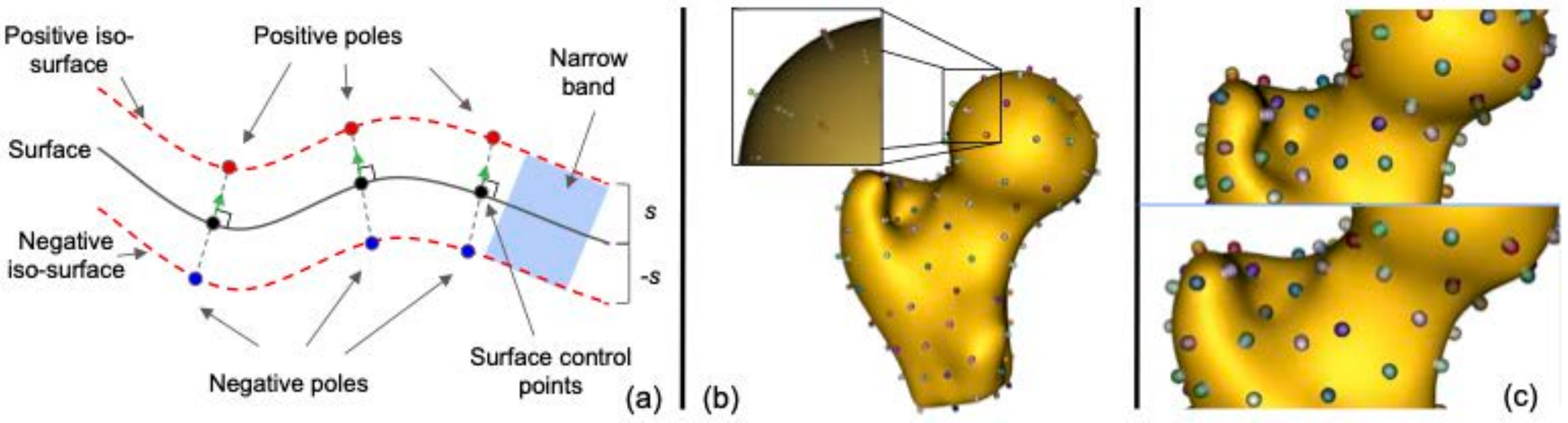}
\caption{(a) Concept of populating a surface using control points and the iso-surfaces using positive and negative pole points. (b) Same concept applied to an output three-dimensional reconstructed femur. (c) Normals can be used to describe very distinct features of the greater trochanter. } \label{dipoles}
\end{figure}

Implicit surface representation based on radial basis functions, \textit{\srepresentation}\ for short, has been proven effective at representing intricate shapes by leveraging both surface control points and normals to inform shape reconstructions \cite{Turk1999VariationalIS, rbf_surface1}. It defines a set of control points at the zero-level set and a pair of off-surface points (aka \textit{dipoles}) with a signed distance $s$ and $-s$ along the surface normal of each control point.
%
This is illustrated in Figure \ref{dipoles}. We refer to the set of control points and their dipoles as $\widetilde{\P_i}$ for shape $i$, where $\widetilde{\P_i} = [\P_i, \P_i^+, \P_i^-]$ with $\p_{i,j}^{\pm} = \p_{i,j} \pm s\n_{i,j}$.
%
%
Using $\widetilde{\P_i}$, we define the shape's implicit function, a function that can query a distance to the surface given a point $\x \in \mathbb{R}^3$, as follows:


\begin{equation} \label{eq:rbf}
\begin{split}
    f_{\widetilde{\P_i}, \w_i}(\x) = \sum_{j \in \widetilde{\P_i}} w_{i,j} \phi(\x, \widetilde{\p}_{i,j}) + \mathbf{c}^T_i \x + c^0_{i} 
\end{split}
\end{equation}

\noindent where $\phi$ is the chosen RBF basis function (e.g., the thin plate spline $\phi(\x,\y) =(|\x-\y|_2)^2\log(|\x-\y|_2)$, the biharmonic  $\phi(\x,\y) =|\x-\y|_2$ or the triharmonic $\phi(\x,\y) =(|\x-\y|_2)^3$) and $\mathbf{c}_i \in \mathbb{R}^3$ and $c^0_{i} \in \mathbb{R}$ encodes the linear trend of the surface. We obtain $\w_i = [w_{i,1}, w_{i,2}, ..., w_{i,3J}, c_i^0, c_i^1, c_i^2,c_i^3] \in \mathbb{R}^{3J + 4}$ by solving a system of equations formed by Eq \ref{eq:rbf} over $\x \in \widetilde{\P_i}$, along with constraints to keep the linear part separate from the nonlinear deformations captured by the RBF term (first term in Eq \ref{eq:rbf}) to form a fully determined system. 
See \cite{Turk1999VariationalIS, rbf_surface1} for more details. Ultimately, we can use this function $f$ to query approximate distances to the surface to build a mesh or a signed distance transform for visualization and analysis.

This representation can better represent shapes with far fewer control points due to its built-in interpolation capabilities, even further enhanced by informing the system with the point normals. Furthermore, this continuous representation allows \ssmmodel\ to adapt to the underlying surface geometry and correct for control point placement mistakes while training.


\subsection{Loss Functions}
\begin{figure}[!t]
\includegraphics[width=\textwidth]{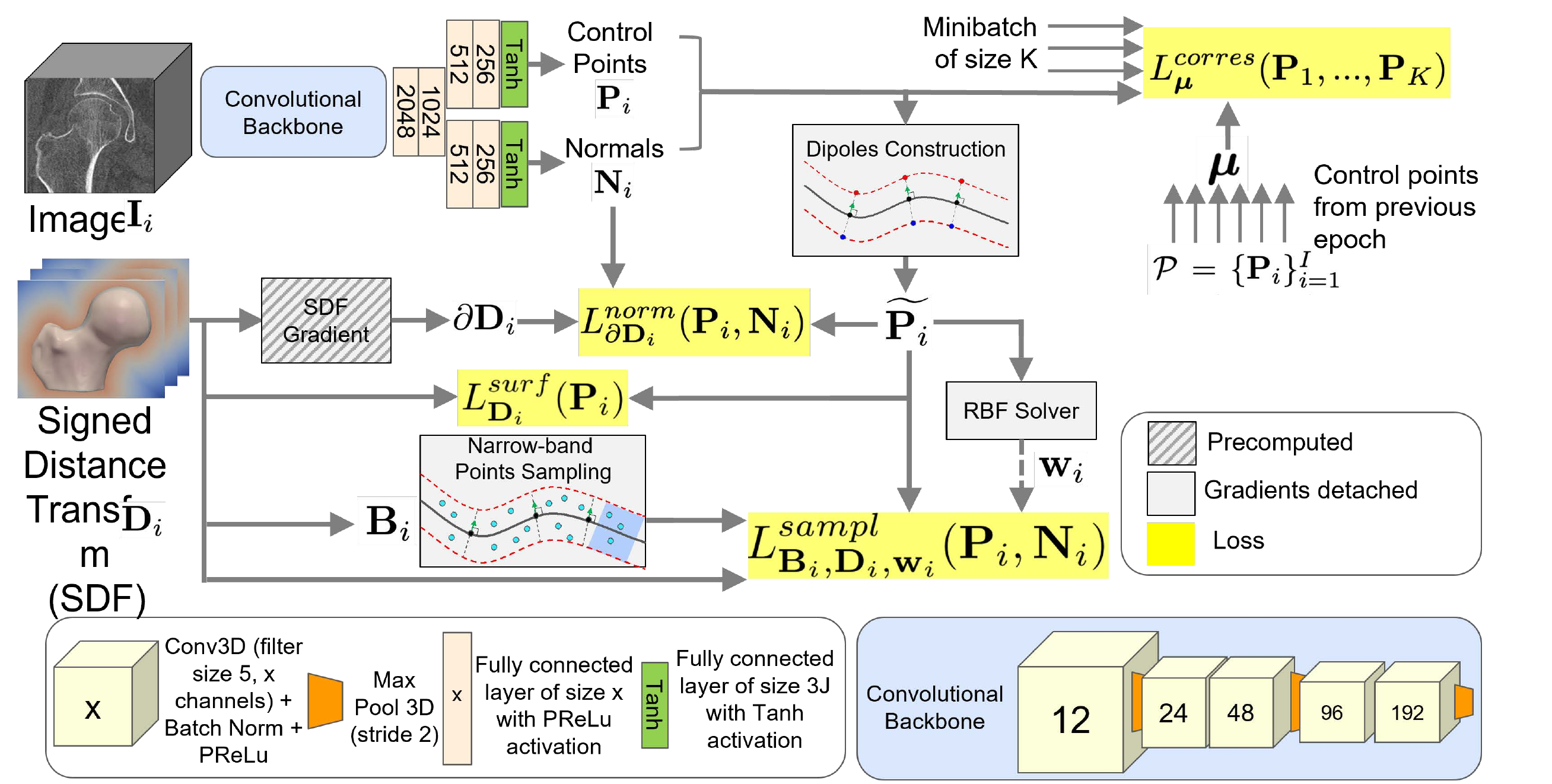}
\caption{The \ssmmodel\ architecture. A 3D image is fed to the convolutional backbone, which produces a flattened output for the feature extractor to produce control points and their respective normals. These are then used to compute the losses of the network.} \label{arch}
\end{figure}

\ssmmodel\ uses four complementary loss functions to be trained on concurrently, illustrated in Figure \ref{arch}. These are (i) \textit{surface loss}, which aims to promote control point and normal adherence to the shape surface, (ii) \textit{normal loss}, which attempts to learn the correct normals at each control point, (iii) \textit{correspondence loss}, which aims to enforce positional correspondence across shapes, and (iv) \textit{sampling loss}, which promotes a spread of the control points that best describes the underlying surface. 
%
%


\vspace{0.1in}
\noindent \textbf{Surface loss:} This loss guides control points to lie on the surface. We use $l^1$-norm to force control points to lie on the zero-level set of the distance transform $\D_i$ by minimizing the absolute distance-to-the-surface evaluated at it. For the $i-$th shape, this loss is defined as,


\begin{equation}
\begin{split}
  L^{surf}_{\D_i} (\P_i) &= \sum_{j = 1}^J  |\D_i(\p_{i,j})|,
\end{split}
\end{equation}

\noindent where $\D_i(\p_{i,j})$ is the distance transform value at point $\p_{i,j}$. 


\vspace{0.1in}
\noindent \textbf{Normal loss:} This loss aims to estimate the surface normal of each control point.  This loss is supervised by the gradient of signed distance transforms (SDF) $\aD = \{\D_i\}_{i=1}^I $, computed from the binary segmentations $\mathcal{S}$, with respect to $\x$, $\partial \aD = \{\partial \D_i\}_{i=1}^I$, which captures unnormalized surface normals. We use the cosine distance (in degrees) to penalize the deviation of the estimated normals from the normals computed from the distance transforms.

\begin{equation}
\begin{split}
  L^{norm}_{\partial \D_i} (\P_i, \bN_i) &= \frac{180}{\pi}  \sum_{j = 1}^J  \cos^{-1} \left(1 - \frac{\n_{i,j}^T \partial\D_i(\p_{i,j})}{\|\n_{i,j}\| \|\partial\D_i(\p_{i,j})\|} \right). 
\end{split}
\end{equation}



\vspace{0.1in}
\noindent \textbf{Correspondence loss:} The notion of control points correspondence across the shape population can be quantified by the information content of the probability distribution induced by these control points in the shape space, the vector space defined by the shapes' PDMs \cite{cates2017shapeworks, cates2007shape}. 
%
The correspondence loss is triggered starting from the second epoch, where the mean shape $\boldsymbol{\mu} = \sum_{i=1}^I \P_i$ is allowed to lag behind the update of the control points. Given a minibatch of size $K$, the correspondence loss is formulated using the differential entropy $H$ of the samples in the minibatch, assuming a Gaussian distribution.

\begin{equation}
\begin{split}
  L^{corres}_{\boldsymbol{\mu}} (\P_1, ..., \P_K) & = H(\P)  
  = \frac{1}{2} \log \left| \frac{1}{3JK}\sum_{k=1}^K \left(\P_k - \boldsymbol{\mu}\right) \left(\P_k - \boldsymbol{\mu}\right)^T \right| ,
\end{split}
\end{equation}
\noindent where $\P$ here indicates the random variable of the shape space.





\vspace{0.1in}
\noindent \textbf{Sampling loss:} This loss makes $f$ encode the signed distance to the surface while encouraging the control points to be adapted to the underlying geometry. Here, we randomly sample $R-$ points $\B_i = [\b_{i,1}, ..., \b_{i,R}]$ that lie within a \textit{narrow band} of thickness $2s$ around the surface (i.e., $\pm s$ from the zero-level set along the surface normal). The sampling loss minimizes distances between these narrow band points and the closest control point to each, scaled by the severity of the distance-to-surface approximation error. This objective guides control points to areas poorly described by $f$ to progressively improve the signed distance-to-surface approximation and represent the shape more accurately.
%
%

Let $\K^i \in \mathbb{R}^{R \times M}$ define the pairwise distances between each narrow band point $\b_{i,r}$ and each control point $\p_{i,j}$ for the $i-$th shape, where its $r,j-$th element $k_{r,j}^i = \|\b_{i,r} - \p_{i,j}\|_2$. Let $\operatorname{softmin}(\K^i)$ encode the normalized (over $\P_i$) spatial proximity of each narrow band point to each control point, where $r$, the $j$th element of $\operatorname{softmin}(\K^i)$ is computed as $\exp{(-k_{r,j}^i)} / \sum_{j'=1}^J \exp (k_{r,j'}^i)$. Let $\e_i \in \mathbb{R}_+^{R}$ captures the RBF approximation squared error at the narrow band points, where $e_{i,r} = [f_{\widetilde{\P_i}, \w_i}(\b_{i,r}) - \D_i(\b_{i,r})]^2$. Let $\E_i = \e_i \mathbf{1}_M^T$, where $\mathbf{1}_M$ is a ones-vector of size $M$. The samples loss can then be written as,

\begin{equation}
\begin{split}
  L^{sampl}_{\B_i, \D_i, \w_i} (\P_i, \bN_i) & = \operatorname{mean} \left(\operatorname{softmin}(\K_i) \otimes \K_i \otimes  \E_i\right) ,
\end{split}
\end{equation}
\noindent where $\otimes$ indicates the Hadamard (elementwise) multiplication of matrices and $\operatorname{mean}$ computes the average over the matrix elements.






\vspace{0.1in}
\noindent \textbf{\ssmmodel\ loss:} Given a minibatch of size $K$, the total loss of \ssmmodel\ can be written as follows:


\begin{equation}
\begin{split}
  L_{\mathcal{I}, \aD, \partial \aD}(\aP_K, \aN_K) &= \sum_{i=1}^K \biggl( \alpha L^{surf}_{\D_i} (\P_i) + 
  \beta L^{norm}_{\partial \D_i} (\P_i, \bN_i) + 
   \gamma L^{sampl}_{\B_i, \D_i, \w_i} (\P_i, \bN_i) \biggl) \\
  & +  \zeta L^{corres}_{\boldsymbol{\mu}} (\P_1, ..., \P_K) 
\end{split}
\end{equation}


\noindent where $\alpha,\beta, \gamma$, and $\zeta \in \mathbb{R}_+$  are weighting hyperparameters of the losses and $\aP_K, \aN_K$ are the control points and normals of the samples in the minibatch. Figure \ref{arch} gives a full overview of the network and its interaction with the loses. \ssmmodel's network is trained end-to-end with $\w_i$s detached from the training so that the loss does not back-propagate through the volatile linear solver. 

%% file: results.tex
\section{Results}



We demonstrate \ssmmodel's performance against the state-of-the-art correspondence optimization algorithm, namely the particle-based shape modeling (PSM), using its open-source implementation, ShapeWorks \cite{cates2017shapeworks}, and DeepSSM \cite{bhalodia2018deepssm, bhalodia2021deepssm}, a deep learning method that trains on an existing correspondence model (provided by the PSM in this case) to infer PDMs on new unsegmented images.



\subsection{Datasets}
We run tests on a dataset consisting of 50 proximal femur CT scans devoid of pathologies in the form of image-segmentation pairs. The femurs are reflected when appropriate and rigidly aligned to a common frame of reference. 
Due to space limitations, we also show similar results for a large-scale left atrium MRI dataset in the supplementary materials. For ease of comparison, we build SSMs with 128 particles for all algorithms as is sufficient to cover important femur shape features (femoral head with its fovea and the lesser and greater trochanter).

\begin{figure}[!t]
\includegraphics[width=\textwidth]{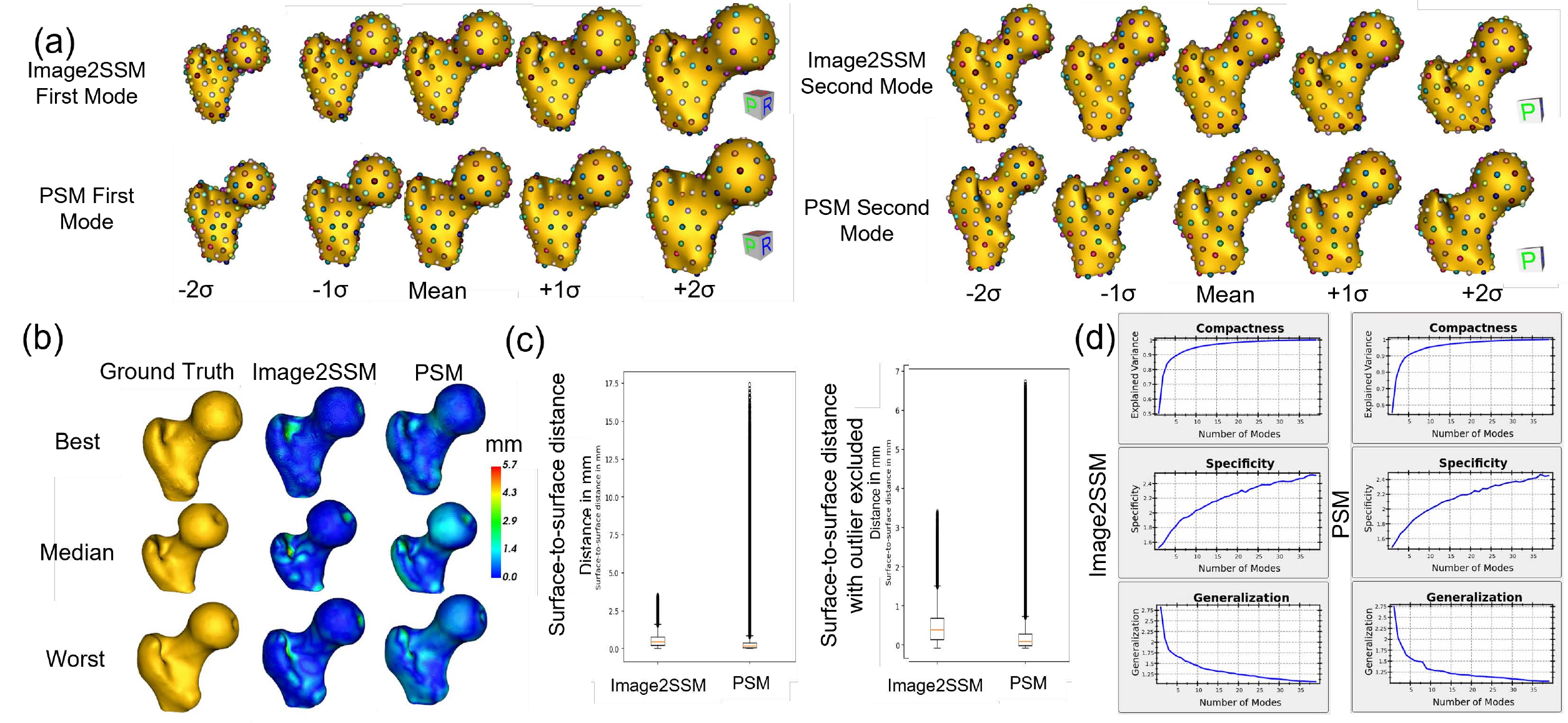}
\caption{(a) First and second modes of variation obtained from \ssmmodel\ training data and PSM. (b) Surface-to-surface distance on a best, median, and worst training femur mesh. (c) The left image shows the surface-to-surface distance comparison on all the data used to train \ssmmodel; the right shows it without outliers.
}  \label{sws}
\end{figure}

\begin{figure}[!t]
\includegraphics[width=\textwidth]{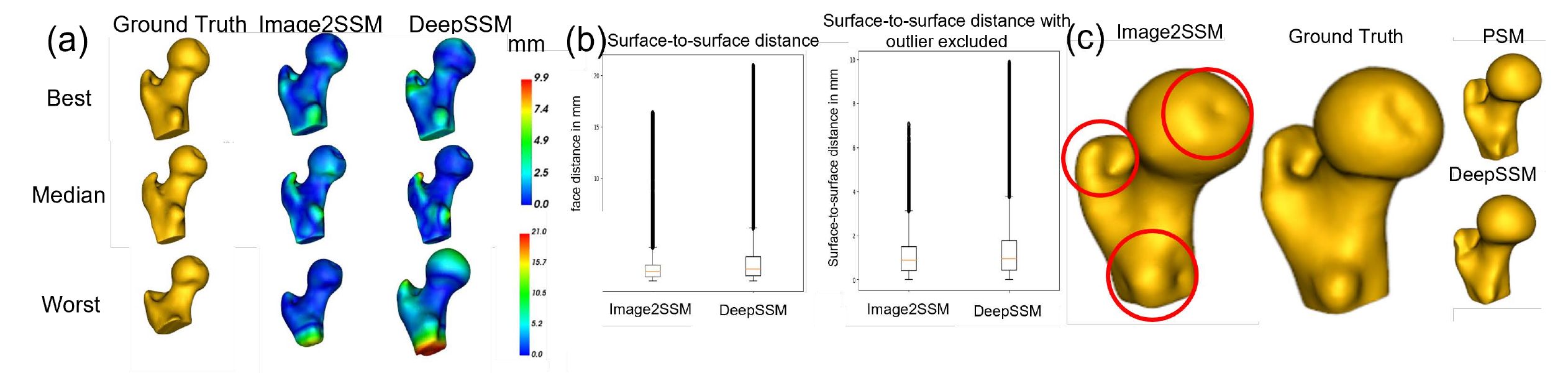}
\caption{(a) Surface-to-surface distance on a reconstructed femur mesh from particles of a few test samples. (b) Surface-to-surface distance plot between DeepSSM and \ssmmodel, and the same plot without the outlier femur. (c) Illustrates \ssmmodel 's capacity to capture detail on an unseen test image. (d) Shows the compactness (higher is better), specificity (lower is better) and generalization (lower is better) graphs against the number of modes of variation. 
} \label{dssm}
\end{figure}

\vspace{0.1in}
\noindent \textbf{Statistical Shape Model: }
We showcase \ssmmodel\ in creating a statistical shape model on its training data and compare such a model with one optimized by PSM \cite{cates2017shapeworks}. 
Figure \ref{sws} showcases the modes of variation, the surface-to-surface distances of \ssmmodel\ against PSM, some representative reconstructions, and graphs for compactness (percentage of variance captured), specificity (ability to generate realistic shapes), and generalization (ability to represent unseen shape instance) \cite{Davies2004LearningSO}. We observe that the modes of variation and metrics match expectations in both approaches. 
We show the effectiveness of \ssmmodel\ in adapting to surface details to achieve lower maximum surface-to-surface distance, and that, unlike PSM, we can achieve reasonable reconstructions using \srepresentation. More on adaptation to detail in figure \ref{dssm}. 

We implement \ssmmodel\ in PyTorch and leverage the Autograd functionality to perform gradient descent using the Adam optimizer \cite{adam}. We randomly sample 10,000 3D points within the narrow band of each surface at each iteration. We use the biharmonic kernel for the basis function. However, the performance of \ssmmodel\ is not significantly influenced by the kernel choice. 
%
%
The hyperparameters we use for \ssmmodel\ are $\alpha = 1e^2,\beta=1e^2, \gamma=1e^4$, and $\zeta=1e^3$ for femurs and $\zeta=1e^6$ for left atria, which were determined based on the validation set. In practice, the runtime of Image2SSM is comparable to PSM for the femurs and roughly 2X faster for the left atria. 
\vspace{0.1in}
\noindent \textbf{Inference Results: }
We compare the inference capabilities of \ssmmodel\ against  DeepSSM on unseen test data. We train DeepSSM with a PDM generated by PSM as supervision. 
For a fair comparison, we use DeepSSM without its augmented data, since \ssmmodel\ does not require augmentation to learn shape models. 
Nevertheless, it is possible to generate and train \ssmmodel\ on augmented data with even more facility than with DeepSSM.
Figure \ref{dssm} shows that \ssmmodel\ compares very favorably to DeepSSM qualitatively and in terms of surface-to-surface distance.

%% file: conclusion.tex

\section{Conclusion}

\ssmmodel\ is a novel deep-learning framework that both builds PDMs from image-segmentation pairs and predicts PDMs from unseen images. It uses an \srepresentation\ able to capture detail by leveraging surface normals at control points, and allows the SSM to adaptively permeate surfaces with high-level detail. \ssmmodel\ represents another step forward in fully end-to-end PDMs and steers the field to utilizing more compact but comprehensive representations to achieve new analytical paradigms. Future directions include removing the requirement that the image-segmentation pairs must be rougly aligned across the cohort and relaxing the Gaussian assumption from correspondence enforcement.

\vspace{0.1in}
\noindent \textbf{Acknowledgment} The National Institutes of Health supported this work under grant numbers NIBIB-U24EB029011. The content is solely the responsibility of the authors and does not necessarily represent the official views of the National Institutes of Health.

%% file: supplementary.tex

\begin{figure}
\includegraphics[width=\textwidth]{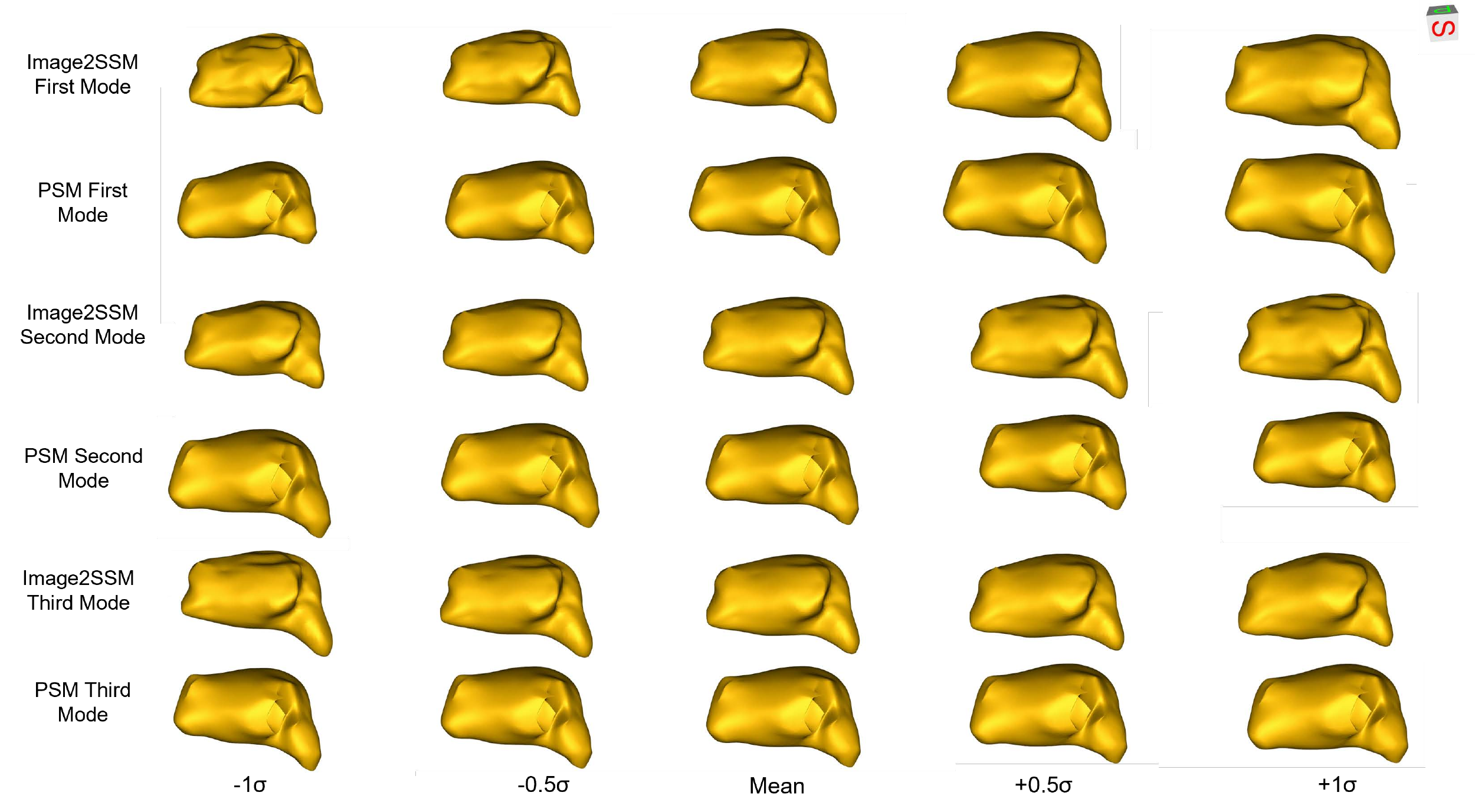}
\caption{We also demonstrate our results on a dataset of 1018 aligned left atrium MRI image-segmentation pairs. This dataset is very challenging due to the high variability in the manual labeling of the pulmonary arteries and the presence of various atrial fibrillation phenotypes (Persistent, paroxysmal, AFL, nonAF, other arrhythmia) As before, we build the model with 128 particles. We show the first three modes of variation of \ssmmodel\ compared to PSM. The results are comparable and match expectations. We observe that both models capture the shape variability of the atrium itself well, less so with the pulmonary arteries. 
}  \label{supp_mov}
\end{figure}

\begin{figure}[!t]
\includegraphics[width=\textwidth]{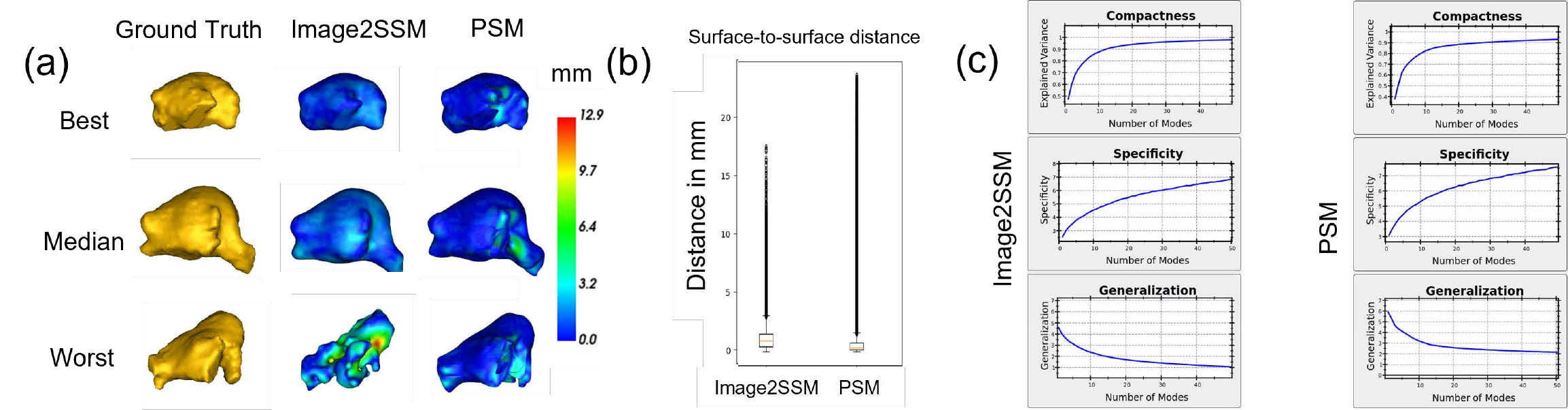}
\caption{(a) Surface-to-surface distance on a best, median, and worst training meshes. (b) The surface-to-surface distance comparison on all the data used to train \ssmmodel. We observe that the distances are comparable between both models in that they capture a large array of shapes well, but fail to different degrees on severe outliers. (c) Shows the compactness (higher is better), specificity (lower is better) and generalization (lower is better) graphs against the number of modes of variation. These are also very similar between the two approaches.
} \label{supp_sws}
\end{figure}

\begin{figure}[!t]
\includegraphics[width=\textwidth]{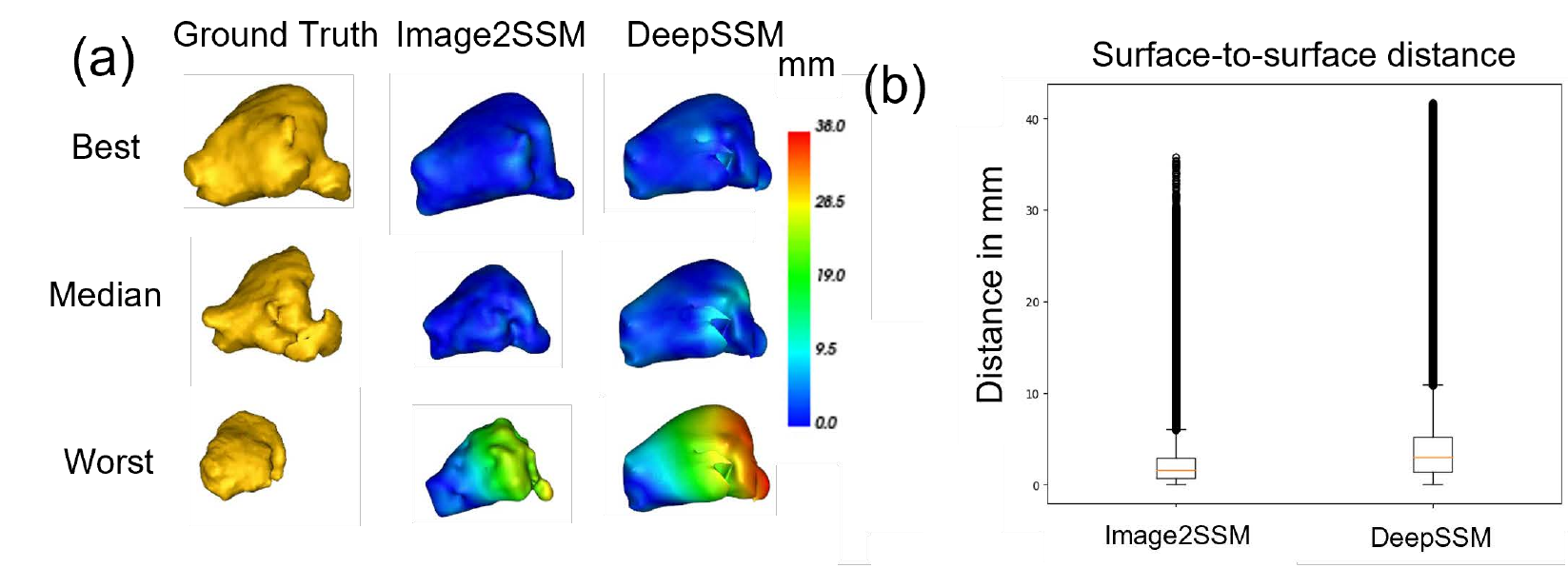}
\caption{(a) Surface-to-surface distance on best, median, and worst held-out samples. (b) Surface-to-surface distance plot between DeepSSM and \ssmmodel. We observe that \ssmmodel\ performs well compared to DeepSSM, but still fails to capture major outliers.
%
%
} \label{supp_dssm}
\end{figure}
